\documentclass[letterpaper, 10 pt, conference]{ieeeconf}  

\usepackage[colorlinks=false,linkcolor=blue,urlcolor=blue,citecolor=blue]{hyperref}
\usepackage{bm}
\usepackage{graphicx}
\usepackage{amsfonts,amssymb,amsmath}
\usepackage{booktabs}
\usepackage{multirow}
\usepackage{stfloats}
\usepackage{caption}
\usepackage{subcaption}

\usepackage[dvipsnames,table,xcdraw]{xcolor}
\usepackage{tcolorbox}

\usepackage[linesnumbered,ruled,vlined]{algorithm2e}
\IEEEoverridecommandlockouts                              

\title{\LARGE \bf
Differentiable Skill Optimisation for Powder Manipulation in Laboratory Automation
}

\author{Minglun Wei$^{1}$, Xintong Yang$^{1}$, Yu-Kun Lai$^{2}$ and Ze Ji$^{1, \dagger}$
\thanks{$^{\dagger}$Corresponding author: Ze Ji. \texttt{JiZ1@cardiff.ac.uk}}
\thanks{$^1$School of Engineering, Cardiff University, Cardiff, CF24 3AA, United Kingdom.}%
\thanks{$^2$School of Computer Science and Informatics, Cardiff University, Cardiff, CF24 4AG, United Kingdom.}%
}

\begin{document}

\maketitle
\thispagestyle{empty}
\pagestyle{empty}

\begin{abstract}
Robotic automation is accelerating scientific discovery by reducing manual effort in laboratory workflows. However, precise manipulation of powders remains challenging, particularly in tasks such as transport that demand accuracy and stability. We propose a trajectory optimisation framework for powder transport in laboratory settings, which integrates differentiable physics simulation for accurate modelling of granular dynamics, low-dimensional skill-space parameterisation to reduce optimisation complexity, and a curriculum-based strategy that progressively refines task competence over long horizons. This formulation enables end-to-end optimisation of contact-rich robot trajectories while maintaining stability and convergence efficiency. Experimental results demonstrate that the proposed method achieves superior task success rates and stability compared to the reinforcement learning baseline.
\end{abstract}

\section{INTRODUCTION} \label{sec:intro}

Robotic automation is accelerating scientific discovery by streamlining workflows in areas such as photocatalysis~\cite{burger2020mobile} and synthetic chemistry~\cite{dai2024autonomous}. These systems reduce manual workload and allow scientists to focus on higher-level reasoning and design~\cite{radulov2025flip}. However, while macroscale processes have seen rapid automation, challenges persist at the microscale, particularly in the precise handling of powders. Powders play a critical role in pharmaceuticals and materials science, where precise transport is essential for reproducibility and system stability. Despite some efforts in weighing~\cite{kadokawa2023learning,radulov2025flip} and grinding~\cite{nakajima2022robotic,nakajima2023robotic}, powder transport is often treated as secondary, and its reliable optimisation remains underexplored.

This research gap is not incidental. The dynamic behavior of powders during motion and interaction is highly nonlinear and sensitive to environmental variability, making it difficult to address with conventional learning or control techniques. 
Recent advances in differentiable physics and high-performance parallel computation~\cite{taichi,difftaichi} present a promising direction for addressing this challenge. By optimising within high-fidelity, real-world-consistent differentiable simulation environments, it becomes possible to obtain precise and reliable powder transport trajectories in a safe manner. While prior studies have applied such methods to the manipulation of materials like elastoplastic solids~\cite{yang2024differentiable, chen2022diffsrl, plb} and fluids~\cite{fluidlab, li2023difffr}, no existing work has systematically addressed the problem of powder transport in laboratory settings using skill or trajectory optimisation.

To address this gap, we propose a trajectory optimisation framework for powder transport in a laboratory setting. Specifically, we build a differentiable physics simulator using Taichi to model powder dynamics and define differentiable skill parameters mapped to control inputs. To tackle the challenges of long-horizon manipulation, we further adopt a curriculum optimisation strategy that first focuses on scooping-related parameters before optimising the full parameter set. These parameters are optimised via gradient backpropagation through a task-specific loss. Comparative experiments with standard reinforcement learning (RL) methods demonstrate the effectiveness and efficiency of our approach.
\begin{figure}[t]
    \centering
    \includegraphics[width=0.9\linewidth]{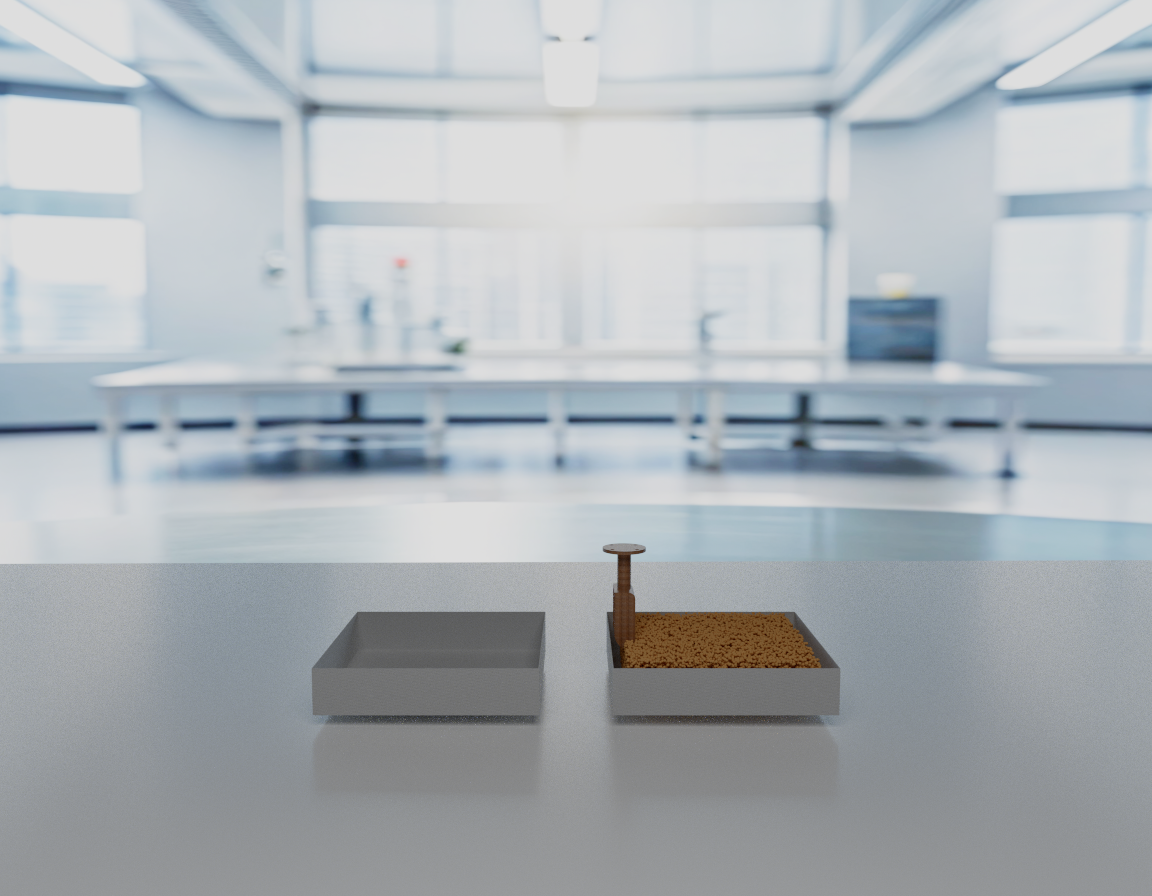}
    \caption{Powder manipulation setup in our simulated environment. Our renderer is built based on LuisaRender~\cite{luisa}.}
    \label{fig:render}
\end{figure}

\section{RELATED WORK}\label{sec:related-work}
\subsection{Powder Manipulation in Lab Automation} \label{sec:rw:pm}
Powder manipulation plays a critical role in lab automation, particularly in pharmaceuticals and materials science, where precise transport is key to maintaining accuracy and material integrity. Prior work has explored powder-related tasks such as grinding~\cite{nakajima2022robotic,nakajima2023robotic}, weighing~\cite{kadokawa2023learning,radulov2025flip}, scooping~\cite{takahashi2025scu}, scraping~\cite{pizzuto2024accelerating}, and dispensing~\cite{jiang2023autonomous}. Notably, \cite{kadokawa2023learning} proposed a sim-to-real RL framework for weighing, while~\cite{radulov2025flip} improved simulation fidelity using flowability-aware Bayesian inference. Although~\cite{takahashi2025scu} tackles scooping, its focus is on hand design rather than control strategy. To date, no existing work systematically addresses trajectory-level optimisation for powder transport in lab settings.
\subsection{Differentiable Simulation-based Manipulation} \label{sec:rw:diff}
Differentiable simulation enables gradient-based reasoning over physical processes, making it a compelling tool for learning control policies. Early examples like DeLaN~\cite{lnn,lnnijrr} applied this to rigid-body systems, though they lack support for complex material interaction. Recent frameworks such as Taichi~\cite{difftaichi} have enabled scalable differentiable simulation across domains like deformable solids~\cite{yang2024differentiable, chen2022diffsrl, plb}, fluids~\cite{fluidlab, li2023difffr}, granular materials~\cite{wei2024automachef}, and thin-shell structures~\cite{wang2024thin}. Among these,~\cite{wei2024automachef} guided RL with differentiable trajectory optimisation in granular media. However, most methods optimise low-level actions per timestep, which can suffer from instability. Our approach instead introduces skill-level parameterisation to improve stability and reduce dimensionality.

\section{METHOD} 
\label{sec:method}
\subsection{Problem Formulation} \label{sec:method:pf}
This study focuses on optimising powder transport trajectories in laboratory settings. We aim to find an optimal set of skill parameters $\Theta$ that generate control sequences of horizon $T$ for the dynamic system. We formulate a time-discretised trajectory optimisation problem with state trajectory $\mathcal{S}= (\mathbf{s}_0, ..., \mathbf{s}_T)$, observations $\mathcal{O}= (\mathbf{o}_0, ..., \mathbf{o}_T)$, and control inputs $\mathcal{U}=(\mathbf{u}_0, ..., \mathbf{u}_{T-1})$. Starting from an initial state $\mathbf{s}_0$, the optimisation is defined as:
\begin{align}
    \min_{\Theta} \ & \mathcal{L}(\mathbf{o}_T, \mathbf{p}^{\text{target}}) \label{eq:1} \\
    \text{s.t.} \ & \mathbf{s}_{i+1} =  \mathbf{f}(\mathbf{s}_i, \mathbf{u}_i)  \label{eq:2} \\
    & \mathbf{u}_i = \mathbf{g}(\Theta)[i] \quad \forall i = 0, \ldots, T{-}1 \label{eq:3} \\
    & \Theta = 
    \begin{cases}
        \Theta^{\text{init}} & j = 0 \\
        \Theta^{-} - \alpha \cdot \nabla_{\Theta^{-}} \mathcal{L}(\mathcal{X}^{-}, \Theta^{-}) & j > 0
    \end{cases} \label{eq:4}
\end{align}
Here, $\mathcal{L}(\cdot)$ measures the distance between the final observation $\mathbf{o}_T$ and the target position $\mathbf{p}^{\text{target}}$, $\mathbf{f}(\cdot)$ represents system dynamics, and $\mathbf{g}(\cdot)$ maps skill parameters to control inputs. The parameter update follows a gradient descent rule with learning rate $\alpha$, using the gradient of the loss evaluated on the previous trajectory and parameters.
\subsection{Task Specification} \label{sec:method:ts}
We consider a powder transport task in a laboratory setting, where the objective is to move as much powder as possible from a source container to an initially empty target container. The simulated environment replicates the real-world configuration: one container is filled with powder, while the other is empty. The system state $\mathbf{s}$ contains full particle-level information (e.g., positions, velocities, accelerations), while the robot has access only to partial observations $\mathbf{o}$, consisting of particle positions. 

The robot action is represented as a 6-dimensional Cartesian displacement $\mathbf{u} \in \mathbb{R}^6$, encompassing translation and rotation. These control inputs are generated from a low-dimensional skill parameter set $\Theta$ via a differentiable mapping $\mathbf{g}(\cdot)$. To guide the optimisation, we define a task-level loss function $\mathcal{L}$ based on spatial proximity, rather than matching a fixed target configuration. Specifically, the loss is computed as the sum of absolute distances between the final observed particle positions $\mathbf{o}_T$ and a designated goal position $\mathbf{p}^{\text{target}}$:
\begin{equation} \label{eq:loss}
\mathcal{L} = \sum_{\mathbf{p}_j \in \mathbf{o}_T} |\mathbf{p}_j - \mathbf{p}^{\text{target}}|
\end{equation}
where $j$ indexes the observed particles. This formulation not only ensures efficient powder delivery to the target while accommodating variations in the final distribution.

\subsection{Skill Parameters}\label{sec:method:sp}
To reduce the complexity of long-horizon 6D control, we introduce a compact representation of robot behaviour through a set of bounded skill parameters $\Theta \in [\theta^{\text{min}}, \theta^{\text{max}}]$. These parameters, derived from human demonstrations, define temporally abstracted behaviours and are mapped to control sequences $\mathcal{U}$ over a time horizon $T$ via a differentiable function $\mathbf{g}(\cdot)$.

\begin{figure}[h]
    \centering
    \includegraphics[width=0.99\linewidth]{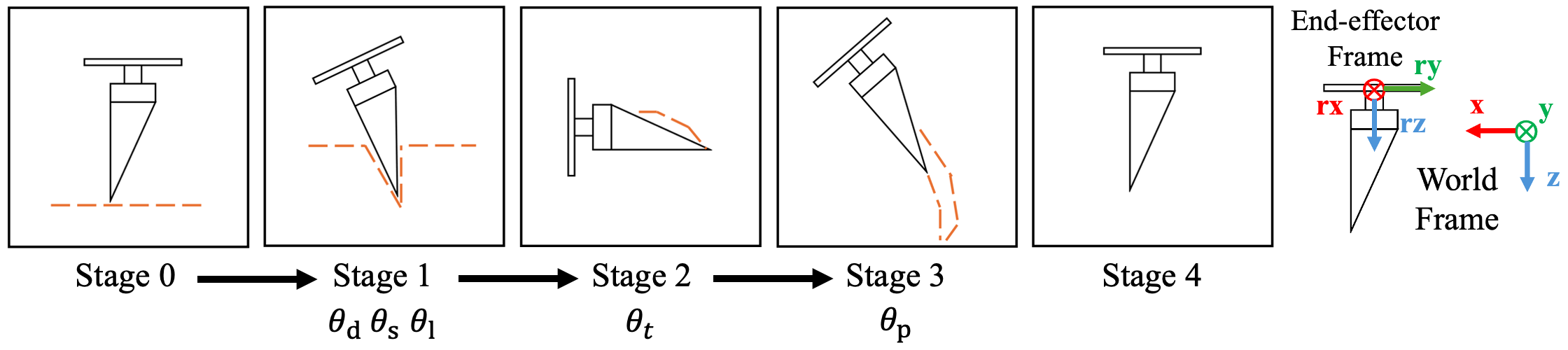}
    \caption{Transport Task Process Visualisation.}
    \label{fig:skill}
\end{figure}

In our implementation, five parameters govern key aspects of the motion: scooping depth $\theta_{\text{d}}$, scooping angle $\theta_{\text{s}}$, post-scoop lifting height $\theta_{\text{l}}$, transport displacement $\theta_{\text{t}}$, and pouring angle $\theta_{\text{p}}$. As shown in Fig.~\ref{fig:skill}, these parameters collectively define the spatiotemporal structure of the robot’s trajectory.

Given fixed translational and rotational velocities, the number of steps for each control segment is computed by dividing the desired displacement in each control dimension by its corresponding velocity and rounding the result. These per-step increments are then distributed across time to form the complete control sequence $\mathcal{U}$.

\subsection{Differentiable Simulation} \label{sec:method:ds}
We simulate granular dynamics using the Moving Least Squares Material Point Method (MLS-MPM)~\cite{mls-mpm}, combined with St. Venant-Kirchhoff elasticity and the Drucker-Prager plasticity model~\cite{Drucker-prager}. Each global time step $\Delta t$ is divided into $N_{\text{sub}}$ substeps of size $\Delta t_{\text{sub}} = \Delta t / N_{\text{sub}}$, with control input scaled as $\mathbf{u}^{\text{sub}} = \mathbf{u} / N_{\text{sub}}$. The global simulation update $\mathbf{f}$ is decomposed into $N_{\text{sub}}$ sub-processes $f$, where each substep is computed as Algorithm~\ref{alg:mpm}.
\begin{algorithm}[h]
    \SetAlgoLined
    \caption{Simplified Sub-step Procedure $f$}
    \label{alg:mpm}     
    
    Grid Reset and Initialisation
    
    $F_{\text{tmp}} = (I + \Delta t_{\text{sub}} C) F$
    
    $U, S, V = f_{\text{svd}}(F_{\text{tmp}})$
    
    $F', \sigma = f_{\text{con}}(U, S, V, \kappa)$
    
    $\mathbf{v}_{\text{grid}} =f_{\text{p2g}}(\mathbf{x}, \mathbf{v}, \sigma, C, \Delta t_{\text{sub}})$

    $\mathbf{s}'_{\text{agent}} = f_{\text{move}}(\mathbf{s}_{\text{agent}}, \mathbf{u}^{\text{sub}}, \Delta t_{\text{sub}})$

    $\mathbf{v}'_{\text{grid}} = \mathbf{v}_{\text{grid}} + \Delta t_{\text{sub}} \mathbf{g}$

    $\mathbf{v}'_{\text{grid}} = f_{\text{col}}(\mathbf{s}'_{\text{agent}}, \mathbf{s}_{\text{con}}, \mathbf{v}'_{\text{grid}}, \Delta t_{\text{sub}})$

    $\mathbf{v}', C' = f_{\text{g2p}}(\mathbf{v}'_{\text{grid}}, \Delta t_{\text{sub}})$

    $\mathbf{v}' = f_{\text{col}}(\mathbf{s}'_{\text{agent}}, \mathbf{s}_{\text{con}}, \mathbf{x},  \mathbf{v}', \Delta t_{\text{sub}})$

    $\mathbf{x}' = \mathbf{x} + \Delta t_{\text{sub}} \mathbf{v}'$

\end{algorithm}

Here, $F$, $C$, $\sigma$, $\mathbf{g}$, and $I$ denote the per-particle deformation gradient, affine velocity field, Cauchy stress, gravity, and identity matrix, respectively. Each $F$ is decomposed as $F = USV^\top$ via singular value decomposition. Material properties are included in $\kappa$. $\mathbf{x}$ and $\mathbf{v}$ are particle positions and velocities, while $\mathbf{s}_{\text{agent}}$ and $\mathbf{s}_{\text{con}}$ denote the robot and container states. Sub-functions include $f_{\text{svd}}$ (SVD), $f_{\text{con}}$ (constitutive model), $f_{\text{p2g}}$ and $f_{\text{g2p}}$ (particle-grid transfers), $f_{\text{move}}$ (robot actuation), and $f_{\text{col}}$ (collision handling). Superscript $'$ indicates updated values at the next substep.

Our framework enables end-to-end optimisation of skill parameters using differentiable physics. Gradients of the task loss are propagated through the entire pipeline, including perception, simulation, and control. All components are differentiable, and gradient flow is carefully maintained across non-smooth operations (e.g., rounding) to ensure end-to-end differentiability.
\subsection{Curriculum Optimisation} \label{sec:method:co}
We formulate powder transport as a long-horizon manipulation task. Prior approaches often divide such tasks into sequential stages such as scooping, transporting, and dumping, with each trained separately and executed in a chained manner~\cite{wei2024automachef}. In contrast, our framework optimises the entire process in an end-to-end fashion, which significantly reduces training overhead. However, both the quantity of material scooped and the accuracy of its deposition are critical to overall task success. In particular, the quality of the initial scooping action directly affects the feasibility of the subsequent transport and pouring stages. To address this, we adopt a curriculum-based optimisation strategy. Specifically, training initially focuses on the first three skill parameters associated with scooping ($\theta_d$, $\theta_s$, $\theta_l$), and gradually expands to include the full parameter set. This progressive approach encourages the agent to acquire reliable foundational behaviours before learning full-sequence coordination, resulting in improved convergence and final performance.
\section{EXPERIMENTS AND RESULTS}
\label{sec:exp}
We compare our trajectory optimisation method with a representative RL baseline, Soft Actor-Critic (SAC)~\cite{sac}, on the powder transport task in a simulated laboratory environment. The simulation replicates real-world conditions, including robot, container, and tool models. Physical parameters are calibrated using our prior DPSI framework~\cite{yang2024differentiable} to ensure sim-to-real consistency. The simulator runs at $\Delta t=0.01$~s with 20 substeps per step.

Skill parameters are optimised within $[-1,1]$ using RMSprop~\cite{rmsprop} with $\beta_r=0.9$ and a learning rate of 0.05, for 30 epochs. In the first 15 epochs, only scoop-related parameters are optimised, and the full parameter set is optimised thereafter. SAC is trained for the same number of episodes, with $\gamma=0.99$, batch size 8, and actor/critic learning rates of 0.001. The task loss (Eq.~\ref{eq:loss}) is negated and used as the reward for SAC to ensure consistent objectives. For evaluation, we report an indicator metric defined as the difference between the target value and the number of particles transported.
\begin{figure}[h]
    \centering
    \includegraphics[width=1.0\linewidth]{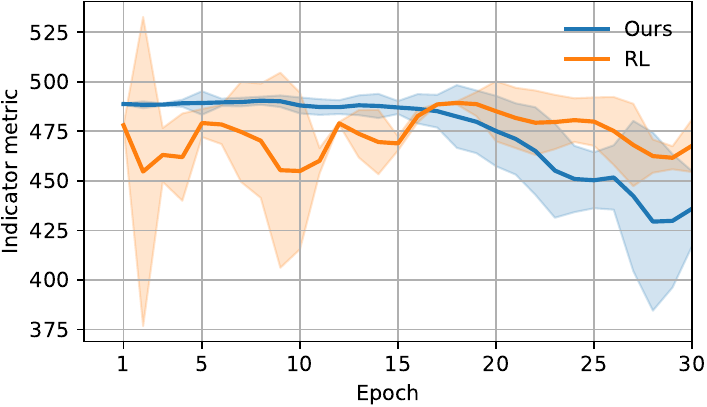}
    \caption{Indicator metric curves for the powder transport task.}
    \label{fig:1}
\end{figure}

To evaluate the performance of our method, we conducted experiments under three different random seeds. The indicator metric curves for our approach and the RL baseline are presented in Fig.~\ref{fig:1}. In the initial stage of training (the first 15 epochs), the indicator metric for our method remains relatively flat. This behaviour is expected, as the full set of parameters is not yet being optimised at this stage. Instead, the robot is primarily acquiring the prerequisite skills required for powder scooping. Once this foundational skill is sufficiently acquired, the optimisation process shifts toward the entire skill set, leading to a more noticeable and stable decrease in indicator metric. Compared to the RL baseline, our method demonstrates a more consistent and eventually better indicator metric trajectory, indicating better convergence and overall performance on the powder transport task.
\section{CONCLUSIONS} \label{sec:con}
This work presents a trajectory optimisation framework for powder transport in laboratory settings. By constructing a differentiable physics simulator, designing differentiable skill mappings, and incorporating a curriculum optimisation strategy, our method addresses the challenges of long-horizon, contact-rich powder manipulation. Experimental results demonstrate that our approach achieves higher task success rates and improved stability compared to the representative RL baseline, underscoring its effectiveness for precision powder handling in laboratory automation scenarios.
\section*{ACKNOWLEDGMENT}
Minglun Wei was supported by the UK Engineering and Physical Sciences Research Council (EPSRC) through a Doctoral Training Partnership (No. EP/W524682/1). This work was also partially supported by the UK EPSRC grant No. EP/X018962/1 and the UK Biotechnology and Biological Sciences Research Council (BBSRC) grant No. BB/Y008537/1.

\bibliographystyle{IEEEtran}
\bibliography{ref} 

\end{document}